\documentclass[pdflatex,sn-mathphys-num]{sn-jnl}

\usepackage[utf8]{inputenc}
\usepackage[T1]{fontenc}
\usepackage{listings}
\usepackage{mathrsfs}

\usepackage{amsmath,amssymb,amsfonts}
\usepackage{graphicx}
\usepackage{bm}
\usepackage{float}
\usepackage{booktabs}
\usepackage{multirow}
\usepackage{array,tabularx}
\usepackage[table]{xcolor}

\usepackage[table]{xcolor}
\definecolor{E3D5CA}{HTML}{E3D5CA}
\definecolor{EDEDE9}{HTML}{EDEDE9}

\usepackage{algorithm}
\usepackage{algpseudocode}

\usepackage{url}
\usepackage{hyperref}
\usepackage[table]{xcolor} 

\definecolor{EDEDE9}{HTML}{EDEDE9}

\usepackage{caption}
\usepackage{subcaption}

\usepackage[title]{appendix}

\theoremstyle{thmstyleone}%
\theoremstyle{thmstyletwo}%

\theoremstyle{thmstylethree}%

\begin{document}

\title[Interactive Nomograms for ICU Survival Prediction]{Development of Interactive Nomograms for Predicting Short-Term Survival in ICU Patients with Aplastic Anemia}


\author[1]{\fnm{Junyi} \sur{Fan}}
\author[1]{\fnm{Shuheng} \sur{Chen}}
\author[1]{\fnm{Li} \sur{Sun}}
\author[1]{\fnm{Yong} \sur{Si}}
\author[2]{\fnm{Elham} \sur{Pishgar}}
\author[3]{\fnm{Kamiar} \sur{Alaei}}
\author[4]{\fnm{Greg} \sur{Placencia}}
\author*[1]{\fnm{Maryam} \sur{Pishgar}}\email{pishgar@usc.edu}

\affil*[1]{\orgdiv{Department of Industrial and Systems Engineering}, \orgname{University of Southern California}, \orgaddress{\city{Los Angeles}, \state{CA}, \country{United States}}}

\affil[2]{\orgdiv{Colorectal Research Center}, \orgname{Iran University of Medical Sciences}, \orgaddress{\city{Tehran}, \country{Iran}}}

\affil[3]{\orgdiv{Department of Health Science}, \orgname{California State University}, \orgaddress{\city{Long Beach}, \state{CA}, \country{United States}}}

\affil[4]{\orgdiv{Department of Industrial and Manufacturing Engineering}, \orgname{California State Polytechnic University}, \orgaddress{\city{Pomona}, \state{CA}, \country{United States}}}


\abstract{Aplastic anemia is a rare, life-threatening hematologic disorder characterized by pancytopenia and bone marrow failure. ICU admission in these patients often signals critical complications or disease progression, making early risk assessment crucial for clinical decision-making and resource allocation. In this study, we used the MIMIC-IV database to identify ICU patients diagnosed with aplastic anemia and extracted clinical features from five domains: demographics, synthetic indicators, laboratory results, comorbidities, and medications. Over 400 variables were reduced to seven key predictors through machine learning-based feature selection. Logistic regression and Cox regression models were constructed to predict 7-, 14-, and 28-day mortality, and their performance was evaluated using AUROC. External validation was conducted using the eICU Collaborative Research Database to assess model generalizability. Among 1,662 included patients, the logistic regression model demonstrated superior performance, with AUROC values of 0.8227, 0.8311, and 0.8298 for 7-, 14-, and 28-day mortality, respectively, compared to the Cox model. External validation yielded AUROCs of 0.7391, 0.7119, and 0.7093. Interactive nomograms were developed based on the logistic regression model to visually estimate individual patient risk. In conclusion, we identified a concise set of seven predictors, led by APS III, to build validated and generalizable nomograms that accurately estimate short-term mortality in ICU patients with aplastic anemia. These tools may aid clinicians in personalized risk stratification and decision-making at the point of care.}

\keywords{Short Term Survival, Aplastic Anemia, Machine Learning, MIMIC-IV Database, Nomogram, Intensive Care Unit}



\maketitle

\section{Introduction}
Aplastic anemia (AA) is a rare but life-threatening hematological disorder characterized by bone marrow failure, leading to reduced hematopoietic stem cells and pancytopenia \cite{young2018aplastic,robinson2018administration}.
The core pathological mechanism of aplastic anemia involves the depletion and dysfunction of hematopoietic stem cells, primarily driven by immune-mediated processes. Studies have shown that cytotoxic T lymphocytes secrete interferon-\(\gamma\) (IFN-\(\gamma\)) and tumor necrosis factor-\(\alpha\) (TNF-\(\alpha\)), which inhibit the proliferation and differentiation of hematopoietic stem cells \cite{young2018aplastic}. In addition, genetic predisposition and environmental factors, such as exposure to certain drugs, chemicals, and viral infections, can further contribute to disruption of the bone marrow microenvironment and exacerbate hematopoietic dysfunction \cite{hoshino2015epidemiology}. 

In Western countries, the incidence of aplastic anemia is approximately 2–3 cases per million people annually, while in East Asian countries, the incidence is higher, reaching 5–7 cases per million \cite{hoshino2015epidemiology,montane2008epidemiology,mary1990epidemiology,norasetthada2021adult}. Without timely and appropriate treatment, aplastic anemia is associated with high mortality rates, primarily due to life-threatening complications such as severe infections, hemorrhage, and multi-organ failure \cite{young2008epidemiology,killick2016guidelines}. In addition to its clinical severity, aplastic anemia profoundly impairs patients' health-related quality of life and imposes a considerable economic and logistical burden on healthcare systems\cite{scheinberg2012treat}.

Treatment of aplastic anemia consists mainly of immunosuppressive therapy (IST) and hematopoietic stem cell transplantation (HSCT). For non-severe cases or patients ineligible for transplantation, IST based on anti-thymocyte globulin (ATG) and cyclosporine is the first-line treatment, achieving an overall response rate of 60--80\% \cite{desmond2018diagnosis}. However, around 30\% of patients may experience relapse or refractory disease \cite{shen2024state}. For severe cases eligible for HSCT, it remains the only potentially curative option, with a five-year survival rate exceeding 80\% when a matched donor is available \cite{desmond2018diagnosis}. In recent years, the introduction of hematopoietic stimulating agents, such as eltrombopag, and immune modulators, has improved outcomes for patients with aplastic anemia, although effective strategies for refractory cases remain an ongoing area of research \cite{townsley2017eltrombopag}. Thus, interpretable and robust prediction of short-term survival in patients with aplastic anemia admitted to the intensive care unit is essential for improving clinical outcomes, as it enables clinicians to tailor therapeutic strategies to individual patient needs, make timely and evidence-based intervention decisions, and allocate critical care resources more effectively.

Scheinberg et al. (2011)\cite{scheinberg2011horse} evaluated the efficacy of horse antithymocyte globulin (ATG) versus rabbit ATG, each combined with cyclosporine, in 120 patients with severe aplastic anemia. The findings demonstrated that early hematologic response within 3 to 6 months to horse ATG was a strong predictor of long-term survival, with a 5-year survival rate of 89\% among responders compared to 60\% in non-responders. Moreover, non-responders exhibited a significantly increased risk of mortality, with a hazard ratio of 3.5 (95\% CI: 1.8–6.7).

Wang et al. (2020)\cite{wang2020pre} examined telomere length (TL) as a biomarker in transplant outcomes. In patients receiving unrelated-donor HSCT for severe AA, pre-transplant leukocyte TL below the 10th percentile (short telomeres) was linked to significantly higher post-transplant mortality (HR = 1.78, 95\% CI 1.18–2.69, P = 0.006) compared to patients with longer telomeres. Thus, short recipient TL was an independent predictor of poor survival, highlighting telomere attrition as a prognostic biomarker in AA.

Liu et al. (2023)\cite{liu2023development} investigated the stratification of risk of early death in very severe AA (VSAA). Using a retrospective cohort of 377 patients with VSAA receiving first-line immunosuppression, they identified key predictors (\( age > 24 \),  $ANC \leq 0.015×10^9/L$, \( ferritin >900 ng/mL \), recurrent fevers) and assigned risk scores. The resulting Early Death Risk Score predicted early mortality well, with an AUROC of 0.862 in the validation set. High-risk patients had significantly higher early death rates, suggesting that the score can guide early transplant decisions.

Although traditional studies on clinical predictors have provided valuable insights into the survival of patients with aplastic anemia, they come with notable limitations. These approaches often depend on a narrow range of variables, which may not adequately reflect the complexity of individual patient conditions. Moreover, they may have difficulty handling high-dimensional data sets or missing information, which can affect their generalizability and predictive performance. Therefore, there is a critical need for more advanced methods that can overcome these challenges and offer accurate and reliable predictions of survival outcomes in patients with aplastic anemia.

In recent years, substantial progress has been made in the development of predictive models for clinical outcomes. Among these, the nomogram has gained increasing attention as a practical and interpretable tool in clinical decision-making. Nomograms offer several key advantages: (1) Nomograms incorporate multivariate regression analysis to provide individualized risk estimates, (2) Nomograms present results in a visual format that enhances clinical interpretability and usability, (3) Nomograms allow for integration of diverse clinical predictors while maintaining interpretability, which is essential for guiding treatment decisions in practice.

This study introduces several methodological innovations that improve the predictive accuracy and clinical utility of machine learning models for short-term survival in ICU patients with aplastic anemia.
\begin{itemize}
\item Three nomograms were developed to predict survival probabilities of 7 days, 14 days and 28 days in ICU patients with aplastic anemia. Each nomogram was constructed based on logistic regression models that incorporate clinically relevant predictors, ensuring both robust statistical performance and medical interpretability. An interactive nomogram application was also created to facilitate real-time individualized survival predictions based on patient-specific variables. This tool improves clinical usability by translating complex modeling outputs into an intuitive graphical interface. 
\item The interactive platform supports personalized management strategies and clinical decision-making by enabling dynamic bedside risk estimation. Together, static and interactive nomograms offer a practical and interpretable solution to improve outcome prediction in critically ill patients with aplastic anemia, bridging advanced analytics with real-world ICU practice.
\item A two-step feature selection framework was implemented to refine over 400 candidate variables to 7 critical predictors. This hybrid approach combined filter-based selection (SelectKBest with F-test scoring) and wrapper-based Recursive Feature Elimination (RFE), augmented by clinical expertise to ensure biological relevance and interpretability.
\item A novel custom SMOTE algorithm was developed specifically for continuous target variables, addressing class imbalance through weighted sampling and synthetic data generation. Thresholds and noise injection were strategically applied to balance mortality predictions in 7-day, 14-day, and 28-day intervals.
\item A comprehensive comparison with traditional models revealed the superiority of logistic regression (LR) over Cox regression and advanced machine learning methods. The LR model achieved AUROC values of 0.8227 (7 days), 0.8311 (14 days), and 0.8298 (28 days), with external validation in the eICU database confirming generalizability (AUROC: 0.7391–0.7093). A Cox-based model was used is also used  to verify our characteristics as in the reference paper, and the C index increased by 18.3\%, demonstrating the effectiveness of the model.
\item The interpretability of the model was enhanced through dual analytical frameworks. SHAP analysis identified time-dependent predictor contributions (e.g., APS III dominance in longer-term predictions), while permutation importance quantified feature impacts. Key variables such as platelet count, bicarbonate, and APS III were validated as critical prognostic markers throughout all time horizons.
\end{itemize}

\section{Methodology}\label{sec:Methodology}

\subsection{Data Source}\label{subsec:Data Source}
The Medical Information Mart for Intensive Care IV (MIMIC-IV) is a comprehensive and freely accessible database that compiles deidentified health data from patients admitted to intensive care units (ICU) at Beth Israel Deaconess Medical Center between 2008 and 2019. This resource covers a wide range of information, including demographics of the patient, vital signs, laboratory results, medications, and clinical notes. The primary purpose of MIMIC-IV is to facilitate critical care research and support the development of predictive models by providing a rich data set that reflects real-world clinical scenarios \cite{johnson2023mimic}.

By integrating data from these comprehensive datasets, we can capture detailed clinical measurements and observations across a heterogeneous patient population. This diversity ensures that the data are representative of various demographics and conditions of the patients, thereby improving the generalizability of the research findings. For studies focusing on rare diseases such as aplastic anemia, MIMIC-IV offers valuable information by providing access to pertinent patient records, which can be instrumental in understanding disease patterns, treatment responses, and outcomes \cite{zhang2024nomogram}.

For external validation, we apply the e-ICU. The Collaborative Research Database eICU is a multicenter critical care database comprising deidentified health data associated with more than 200,000 admissions to the ICU in the United States \cite{pollard2018eicu}.

\subsection{Study Population and study design}\label{subsec:Study Population}
The population selection focuses on ICU patients diagnosed with aplastic anemia or related conditions using ICD-9 codes (\textbf{284}) and ICD-10 codes (\textbf{D60}, \textbf{D61}).
 Patients are filtered to include only those with an ICU stay of at least one day and who survived beyond the first 24 hours of ICU admission. Additionally, only patients aged between 18 and 90 years are included. For patients with multiple stays in the ICU, only the first admission is considered, ensuring one record per subject. This approach ensures a clinically meaningful cohort by excluding short-term stays, early deaths, and patients outside the target age range. The final data set includes subject identifications, hospital admissions, stays in the ICU and timestamps of admission, discharge, and death for further analysis.
The extraction workflow for the study population is shown in \textbf{Figure 1.}

\begin{figure}[H]
\centering
\includegraphics[width=0.75\linewidth]{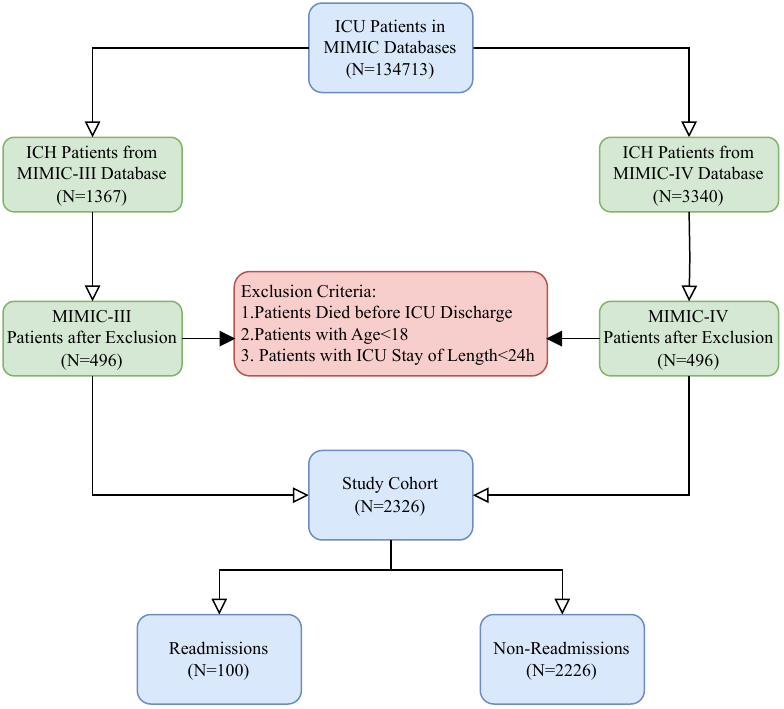}
\captionsetup{justification=raggedright,singlelinecheck=false}
\caption{Criterion of study population extraction}
\label{population}
\end{figure}

This study proposes a structured machine learning pipeline to predict short-term mortality (7, 14, and 28 day) in ICU patients with aplastic anemia. Patients were identified from the MIMIC-IV database using ICD-9/10 codes, with inclusion criteria ensuring ICU stays over 24 hours, age between 18--90, and one admission per patient.

Over 400 clinical variables were extracted across five domains. Features with $>$80\% missingness were removed, followed by KNN imputation and z-score normalization. Feature selection combined SelectKBest and Recursive Feature Elimination, yielding seven predictors with low multicollinearity (VIF $<$ 5).

To handle outcome imbalance, a custom SMOTE approach was applied using threshold-based weighting and noise-enhanced interpolation. Models including logistic regression, Cox, and tree-based methods were trained with 70/30 splits and evaluated using AUROC and PR-AUC.

Model interpretation was performed using SHAP and permutation importance, highlighting APS III, platelet count, and bicarbonate as key predictors. External validation using the eICU database confirmed model generalizability. An interactive Dash-based nomogram was also developed to support real-time clinical use.

In summary, we implemented a clinically informed machine learning framework for predicting short-term ICU mortality in patients with aplastic anemia. The pipeline integrates cohort selection, high-dimensional feature preprocessing, class imbalance handling, model training, and interpretability analysis.  As shown in Algorithm~\ref{alg:aa_nomogram}, the workflow supports both static and interactive nomogram deployment, enabling individualized risk assessment and potential integration into ICU decision support systems.

\begin{algorithm}[ht]
\caption{\textbf{ML Pipeline for Short-Term Mortality Prediction in ICU Patients with Aplastic Anemia}}
\label{alg:aa_nomogram}
\begin{algorithmic}[1]
\Require MIMIC-IV ICU data with confirmed aplastic anemia diagnosis (ICD-9: 284, ICD-10: D60, D61)
\Ensure Binary predictions: mortality at 7, 14, and 28 days

\State \textbf{Step 1: Cohort Selection} -- Filter by ICD codes; exclude ICU stay $<$ 24h, age $<$18 or $>$90, repeated admissions; retain first ICU stay.

\State \textbf{Step 2: Feature Extraction and Preprocessing} -- Extract $>$400 variables from five domains; remove features with $>$80\% missing; impute using KNN; normalize with z-score; compute derived scores (e.g., APS III).

\State \textbf{Step 3: Feature Selection} -- Apply SelectKBest to select top 50 features; use Recursive Feature Elimination (RFE) to reduce to 7 key predictors; verify multicollinearity via VIF $<$ 5.

\State \textbf{Step 4: Imbalance Correction} -- Use custom SMOTE for continuous target; segment based on thresholds \{7, 35\} with weights \{40, 20, 1\}; interpolate synthetic samples with noise $\epsilon \sim \mathcal{U}(-\delta, \delta)$; generate 2000 new samples.

\State \textbf{Step 5: Model Development} -- Use 70/30 stratified train-test split; evaluate models \{LR, Cox, RF, XGBoost, LightGBM, AdaBoost, NN\}; perform GridSearchCV; metrics: AUROC, PR AUC, sensitivity, specificity.

\State \textbf{Step 6: Interpretation and Validation} -- Apply SHAP and permutation importance for feature attribution; visualize calibration curves and AUROC; compare feature stability.

\State \textbf{Step 7: External Validation} -- Test trained LR model on eICU dataset; compare 7, 14, and 28 day AUROCs to internal test set.

\State \textbf{Step 8: Nomogram Deployment} -- Generate static nomograms for each prediction window; implement interactive Dash-based tool for clinical use.

\end{algorithmic}
\end{algorithm}

\subsection{Feature Selection}\label{subsec:Identifying Features}
In this study, the selection of characteristics was carried out in two sequential steps to identify the most relevant predictors to model short-term survival in ICU patients with aplastic anemia. Initially, a filter-based method, SelectKBest, was applied to narrow down the pool of over 400 candidate variables to the top 50 features based on their statistical association with the outcome variable . 

 SelectKBest ranks features based on a chosen statistical metric, such as the F-test for classification tasks:

\begin{equation}
F = \frac{\text{variance between classes}}{\text{variance within classes}}
\end{equation}

where a higher $F$ value indicates stronger relevance to the target variable. Features with the top $K$ scores are retained, effectively removing noise and redundant information. 

By applying SelectKBest, we ensured that only the most statistically significant features contributed to downstream modeling. This step streamlined our predictive framework, enabling better generalization without sacrificing essential clinical information. The reduced feature set facilitated more efficient training and improved model interpretability while preserving predictive performance\cite{brownlee2019selectkbest}.

Subsequently, a wrapper-based method, Recursive Feature Elimination (RFE), was employed to further refine the feature set. Recursive Feature Elimination (RFE) is an iterative feature selection technique that eliminates the least important features based on model performance, optimizing predictive power while reducing dimensionality. At each iteration, a model assigns weights $\beta_i$ to features, and the least significant feature is removed:

\begin{equation}
\hat{y} = \sum_{i=1}^{n} \beta_i x_i
\end{equation}

where $\beta_i$ represents the feature importance. This approach iteratively evaluates subsets of features by training and testing the model performance, ultimately identifying the optimal set of 7 features that contribute the most to predictive accuracy \cite{brownlee2019rfe, Si2025.05.05.25327009}.

At the same time, we take expert advice on the selecting process. The final selected features were: \textit{APS III}, \textit{Anion Gap}, \textit{Base Excess}, \textit{Bicarbonate}, \textit{Fibrinogen, Functional}, \textit{Lactate Dehydrogenase (LD)}, and \textit{ platelet count}. These variables reflect key clinical and biochemical parameters associated with patient outcomes and serve as the foundation for building the predictive nomogram \cite{kurukulasuriya2009serum}.

The selected features are shown in \textbf{Table 1.}

\begin{table}[H]
\caption{Selected clinical features used for mortality prediction.}
\label{tab:clinical_features}
\centering
\setlength{\tabcolsep}{6pt}
\begin{tabular}{ll}
\hline
\rowcolor{EDEDE9}
\textbf{Category} & \textbf{Feature Name} \\
\hline
Synthetic Indicator & APS III (Acute Physiology Score III) \\
\hline
Laboratory Event & Anion Gap \\
 & Base Excess \\
 & Bicarbonate \\
 & Fibrinogen, Functional \\
 & Lactate Dehydrogenase (LD) \\
 & Platelet Count \\
\hline
\end{tabular}
\end{table}

 Among these, APS III is used to gauge the severity of bone marrow failure, providing an early prognostic indicator \cite{apsiii2011}. The APS III score is calculated using 17 physiological variables, each weighted according to its deviation from the normal range. The final score ranges from 0 to 252, where a higher score indicates greater severity of the illness.

Mathematically, APS III is expressed as:
\begin{equation}
    \text{APS III} = \sum_{i=1}^{n} w_i \cdot x_i,
\end{equation}
where $x_i$ represents the observed value of the $i^{th}$ physiological variable, and $w_i$ is the corresponding weight assigned based on the deviation from normal values. The physiological parameters include heart rate, mean arterial pressure, respiratory rate, oxygenation levels (PaO$_2$/FiO$_2$), blood pH, sodium, potassium, glucose, creatinine, blood urea nitrogen, white blood cell count, hematocrit, temperature, urine output and Glasgow Coma Scale (GCS).

The anions gap has also been associated with metabolic stress in aplastic anemia, where elevated levels may reflect underlying complications \cite{AnionGap2012}.

Base excess, which serves as a marker of acid-base balance, has been reported to correlate with disease progression in marrow suppression \cite{BaseExcess2013}, while bicarbonate levels can offer information on body compensatory responses during bone marrow insufficiency \cite{Bicarbonate2014}.

Here is the calculation of base excess:
\begin{equation}
BE = HCO_3^- - 24.4 + (2.3 \times \text{Hb} + 7.7) \times (\text{pH} - 7.4)
\end{equation}

\begin{itemize}
    \item \( HCO_3^- \) : Bicarbonate concentration in blood, measured in milliliters per liter (mEq / L).
    \item \( 24.4 \) : Normal bicarbonate concentration at standard conditions (mEq/L).
    \item \( \text{Hb} \) : Hemoglobin concentration in the blood, measured in grams per deciliter (g/dL).
    \item \( \text{pH} \) : Arterial blood pH, which represents the acid-base status of the patient.
    \item \( 2.3 \) : Coefficient related to the buffering capacity of hemoglobin.
    \item \( 7.7 \) : Constant derived from standard physiological conditions.
    \item \( 7.4 \) : Normal arterial blood pH.
\end{itemize}

Fibrinogen, a key clotting factor, is often monitored to assess the risk of coagulopathy, which can be exacerbated by reduced platelet production \cite{Fibrinogen2016}. Furthermore, lactate dehydrogenase (LD) frequently serves as a marker of tissue turnover, with elevated values suggesting more aggressive disease activity \cite{LD2018}.  Finally, Platelet Count remains a central measure for evaluating both disease severity and treatment response, where chronic thrombocytopenia signals poorer clinical outcomes \cite{Platelet2019}.

\begin{figure}[htbp]
\centering
\includegraphics[width=0.8\linewidth]{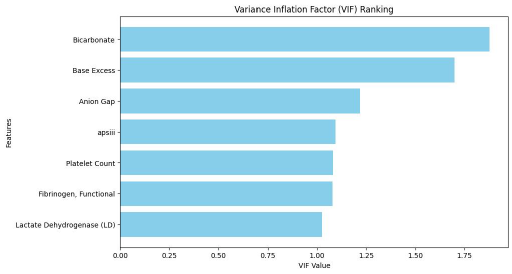}
\caption{Variance inflation factor of selected features.}
\label{fig:vif}
\end{figure}

To assess multicollinearity among the selected features, we computed the variance inflation factor (VIF) for each variable, as illustrated in Figure~\ref{fig:vif}. VIF is a statistical measure that quantifies how much the variance of a regression coefficient is inflated due to collinearity with other predictors. A VIF value greater than 5 is commonly used as a threshold to indicate high multicollinearity, which can compromise model reliability by inflating standard errors and reducing interpretability~\cite{raykov2019vif}.

In this study, all selected predictors demonstrated VIF values well below the commonly accepted threshold of 5, indicating minimal multicollinearity between variables. This confirms that the selected features, including \textit{APS III}, \textit{Anion Gap}, \textit{Base Excess}, \textit{Bicarbonate}, \textit{Fibrinogen, Functional}, \textit{Lactate Dehydrogenase (LD)}, and \textit{ platelet count}, are independent and suitable for use in predictive modeling without significant redundancy.

\subsection{Custom SMOTE for imbalance data}
\label{subsec:custom_smote}

To mitigate the imbalance in the continuous target variable, we developed a customized version of the Synthetic Minority Oversampling Technique (SMOTE), specifically designed for continuous outcome modeling. Unlike standard SMOTE, which is typically applied to categorical data, the proposed method introduces a weighted sampling strategy based on user-defined target thresholds. This design allows for oversampling to be concentrated in underrepresented outcome intervals, enhancing the distributional balance of the training data.

The algorithm begins by segmenting the target variable $Y$ into predefined intervals $\{T_1, T_2, \dots, T_n\}$, each associated with a sampling weight $\{w_1, w_2, \dots, w_{n+1}\}$. Every instance in the original dataset is assigned a sampling probability proportional to the weight of its interval, computed as:
\begin{equation}
    p_i = \frac{w_i}{\sum_{j=1}^{N} w_j}, \quad i = 1, \dots, N
\end{equation}
where $N$ is the number of original samples.

For each sampled instance $y_i$, a synthetic observation is generated by interpolating among $k$ randomly selected nearest neighbors. The new value is computed as:
\begin{equation}
    y_i^{\text{new}} = \frac{1}{k} \sum_{j=1}^{k} y_j + \epsilon
\end{equation}
where $\epsilon \sim U(-\delta, \delta)$ represents a small uniformly distributed noise term added to introduce variability and prevent overfitting.

The augmented dataset is formed by appending the synthesized samples to the original dataset, yielding an expanded training set $(X_{\text{resampled}}, Y_{\text{resampled}})$ that better represents the full range of the target distribution.

In our implementation, we used thresholds $\{7, 35\}$ with associated weights $\{40, 20, 1\}$ and generated 2000 synthetic samples. Table~\ref{tab:dataset_shape} presents the dataset dimensions before and after resampling.

\begin{table}[H]
\caption{Dataset dimensions before and after applying custom SMOTE.}
\label{tab:dataset_shape}
\centering
\setlength{\tabcolsep}{6pt}
\begin{tabular}{lcc}
\hline
\rowcolor{EDEDE9}
\textbf{Dataset} & \textbf{Original Shape} & \textbf{Resampled Shape} \\
\hline
$X$ & $(m, d)$ & $(m + 2000, d)$ \\
$Y$ & $(m, 1)$ & $(m + 2000, 1)$ \\
\hline
\end{tabular}
\end{table}

This method effectively balances the distribution of continuous targets while maintaining statistical coherence, reducing bias towards underrepresented regions in the dataset.

\subsection{Modeling}\label{subsec:Modeling}

In this study, the main objective was to develop robust predictive models for short-term survival in ICU patients with aplastic anemia. Two main models were applied to construct interpretable \textit{nomograms}: Logistic Regression (LR) Logistic Regression is a widely used statistical model for binary classification tasks, such as predicting survival (yes/no) within a specific time period. Estimates the probability of an event occurring based on linear combinations of independent variables. The simplicity and interpretability of LR make it a favored choice for constructing nomograms, where the model coefficients are used to assign scores for individual risk assessment.

Logistic regression probability estimation model:

\begin{equation}
P(Y=1 \mid X) = \frac{1}{1 + e^{-(\beta_0 + \sum_{i=1}^{n} \beta_i x_i)}}
\label{eq:logistic_general}
\end{equation}

To predict 7-day, 14-day, and 28-day mortality of ICU patients with aplastic anemia, we use the following models:

\begin{equation}
P_{7d} = \frac{1}{1 + e^{-(\beta_{0,7} + \sum_{i=1}^{n} \beta_{i,7} x_i)}}
\label{eq:logistic_7d}
\end{equation}

\begin{equation}
P_{14d} = \frac{1}{1 + e^{-(\beta_{0,14} + \sum_{i=1}^{n} \beta_{i,14} x_i)}}
\label{eq:logistic_14d}
\end{equation}

\begin{equation}
P_{28d} = \frac{1}{1 + e^{-(\beta_{0,28} + \sum_{i=1}^{n} \beta_{i,28} x_i)}}
\label{eq:logistic_28d}
\end{equation}

where \( x_i \) represents the input features:

Cox regression, on the other hand, is a semiparametric model commonly used in survival analysis. Estimates the hazard ratio for a given set of covariates without making strong assumptions about the underlying hazard function. This model allows for the prediction of event probabilities over time, making it highly suitable for creating time-dependent nomograms. In this study, Cox regression was used to predict survival probabilities of 7 days, 14 days, and 28 days.

To ensure the robustness and feasibility of the selected features, as well as to benchmark against traditional models, several machine learning algorithms were employed. These include Random Forest (RF), XGBoost (Extreme Gradient Boosting), LightGBM, Neural Networks (NN), and AdaBoost. These models were selected to represent a diverse set of learning paradigms, including ensemble methods, boosting techniques, and deep learning architectures.

Random Forest, a bagging-based ensemble method, is robust to overfitting and capable of capturing non-linear relationships and feature interactions. XGBoost and LightGBM are gradient boosting frameworks known for their high predictive accuracy, computational efficiency, and built-in handling of missing data. Both are particularly effective in structured data tasks and are widely used in clinical prediction challenges. AdaBoost offers an alternative boosting strategy by sequentially correcting misclassified instances and improving weak learners.

Neural networks were also explored to evaluate the ability of non-linear function approximation in capturing complex feature interactions that may not be apparent in traditional models. Although less interpretable, neural networks serve as a valuable comparative benchmark, especially when evaluated alongside explainability methods such as SHAP.

Collectively, these machine learning models served to validate the effectiveness of the selected features and the reliability of the logistic and Cox regression models. Their inclusion ensures that the final modeling framework is not only interpretable but also competitive in predictive performance across multiple algorithmic paradigms.

\section{Results}
\subsection{Statistical Comparison}\label{subsec:StatisticalComparison}

A statistical comparison between the training and test data sets was performed to assess the distribution of key clinical variables and verify the suitability of the cohort division. Table~\ref{table:model_performance_comparison} summarizes the mean and standard deviation (SD) for each variable in both sets, together with the corresponding p-values derived from independent two-sample t-tests.

\begin{table}[htbp]
\caption{Comparison of population features between training and test datasets.}
\label{table:model_performance_comparison}
\centering
\setlength{\tabcolsep}{8pt}
\begin{tabular}{lllll}
\hline
\rowcolor{EDEDE9}
\textbf{Variable} & \textbf{Unit} & \textbf{Train (SD)} & \textbf{Test (SD)} & \textbf{P-value} \\
\hline
APS III & – & 51.8 (18.5) & 52.9 (18.8) & 0.273 \\
Anion Gap & mEq/L & 12.9 (2.6) & 12.9 (2.6) & 0.913 \\
Base Excess & mEq/L & -1.1 (3.5) & -1.2 (3.4) & 0.712 \\
Bicarbonate & mEq/L & 23.8 (3.2) & 23.6 (3.3) & 0.311 \\
Fibrinogen & mg/dL & 337.4 (143.9) & 345.2 (147.8) & 0.324 \\
LD & U/L & 283.1 (112.9) & 291.6 (119.1) & 0.177 \\
\rowcolor{E3D5CA}
Platelet Count & $\times 10^9$/L & 95.8 (54.4) & 89.9 (51.5) & \textbf{0.037} \\
\hline
\end{tabular}
\end{table}

APS III, a widely used composite severity score in ICU risk stratification, did not show significant differences between training (mean = 51.8, SD = 18.5) and test sets (mean = 52.9, SD = 18.8, p = 0.273). Similarly, critical metabolic parameters such as adenosine equilibria, excess base, and bicarbonate levels remained statistically comparable between groups (p-values = 0.913, 0.712, and 0.311, respectively). The consistency of these markers indicates that baseline systemic status and metabolic conditions were similar in the two datasets.

Coagulation-related parameters, including fibrinogen and platelet count, were also examined. Although fibrinogen levels did not show significant differences (p = 0.324), platelet count exhibited a marginally significant variation (mean = 95.8 vs 89.9, p = 0.037). Although statistically significant, the magnitude of the difference in platelet counts was relatively small and is unlikely to materially impact the generalization of the overall model. Furthermore, lactate dehydrogenase (LD), a biomarker indicative of cellular injury and systemic inflammation, did not differ significantly (p = 0.177), supporting the idea that the degree of underlying tissue damage was comparable between the groups.

In general, the distributional similarity of the majority of clinical variables suggests that the training and test cohorts are well balanced, minimizing the risk of selection bias. This is crucial for ensuring the external validity and robustness of the machine learning models, as it implies that learned patterns are likely to generalize to unseen patient populations rather than being artifacts of specific data splits. Furthermore, maintaining clinical consistency across cohorts improves the interpretability of models by providing a stable clinical context in which model output can be meaningfully related to real-world patient physiology. In particular, the alignment of key physiological, metabolic and coagulation markers ensures that any differences in model performance are attributable to true predictive learning rather than confounding cohort discrepancies.

\subsection{Model Performance}\label{subsec:Model Performance}

\begin{figure}[htbp]
\centering
\includegraphics[width=85mm]{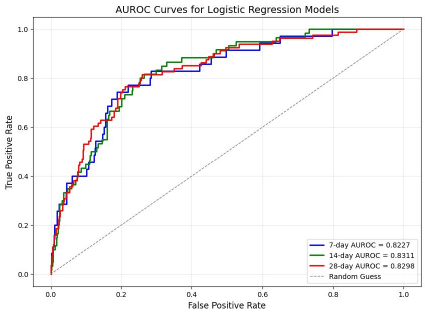}
\caption{AUROC-curves for test set of LR Models}
\label{fig:auroc_logr}
\end{figure}

The Logistic Regression (LR) model demonstrated consistent and reliable predictive performance at all assessed time points. For the prediction of survival at 7 days, the model achieved an AUROC of 0.823 (95\% CI: 0.737--0.909), indicating strong discrimination between survivors and non-survivors. Similarly, for 14-day and 28-day survival predictions, the AUROC values were 0.831 (95\% CI: 0.766--0.896) and 0.830 (95\% CI: 0.773--0.887), respectively, confirming the stable performance of the model over varying clinical time horizons. Although Precision-Recall AUC values (PR AUC) increased from 0.323 at 7 days to 0.516 at 28 days, reflecting improved positive predictive capability in longer-term survival predictions, the LR model consistently provided reliable estimates in different temporal settings. The AUROC curves for our proposed model are illustrated in Fig~\ref{fig:auroc_logr}.

Although several machine learning models, including Random Forest (RF), Neural Networks (NN), LightGBM, and XGBoost, were evaluated to predict mortality at time points of 7 days, 14 days, and 28 days, their overall performance was found to be inferior to the Logistic Regression (LR) model. Despite achieving high specificity, these models consistently demonstrated zero sensitivity at all time points, reflecting the challenges of correctly predicting positive cases. LR outperformed these models in both AUROC and sensitivity, highlighting its robustness and suitability to develop reliable nomograms for the prediction of short-term mortality in ICU patients with aplastic anemia. Other models are shown in Table\ref{table:reg_model_auroc} strong numerical performance, the Logistic Regression model offers substantial advantages in clinical interpretability. By providing direct, interpretable coefficients for each characteristic, LR models allow clinicians to understand the impact of each variable on survival risk, facilitating informed decision making. The clear association between clinical predictors and outcomes enhances the transparency of the model and supports its integration into bedside practice, particularly through the development of easy-to-use nomograms for individualized patient risk assessment.

Although several alternative machine learning models, including Random Forest (RF), Neural Networks (NN), LightGBM, and XGBoost, were evaluated for mortality prediction at 7, 14, and 28 days, their overall performance was found to be inferior to the LR model. Despite achieving high specificity, these models consistently demonstrated zero sensitivity across all time points, reflecting the difficulties in correctly identifying true positive cases. This limitation severely hampers their clinical applicability, particularly in critical care, where early detection of high-risk patients is paramount. LR not only outperformed these models in AUROC, but also maintained reasonable sensitivity levels, highlighting its superior robustness in the face of imbalanced data and limited sample sizes. The performance of other models is detailed in Table~\ref{table:reg_model_auroc}.

\begin{table}[htbp]
\caption{Model performance comparison based on AUROC at different time points.}
\label{table:reg_model_auroc}
\centering
\setlength{\tabcolsep}{8pt}
\begin{tabular}{lccc}
\hline
\rowcolor{EDEDE9}
\textbf{Model} & \textbf{7-day AUROC} & \textbf{14-day AUROC} & \textbf{28-day AUROC} \\
\hline
\rowcolor{E3D5CA}
\textbf{RF} & \textbf{0.8324} & \textbf{0.8138} & \textbf{0.7960} \\
XGBoost & 0.7400 & 0.7288 & 0.7188 \\
LightGBM & 0.7883 & 0.7677 & 0.7527 \\
NN & 0.8242 & 0.7957 & 0.7951 \\
\hline
\end{tabular}
\end{table}

Our Cox regression model, developed alongside LR, significantly outperformed the reference study by Tu et al.~\cite{tu2024nomogram}, achieving a C-index of 0.761 compared to 0.643, representing an improvement of 18. 3\%. This advancement can be attributed to our refined feature selection process, which leveraged machine learning techniques to identify the most informative predictors, as well as comprehensive data preprocessing strategies, including KNN imputation for missing data and rigorous outlier management. These methodologies not only improved the integrity and robustness of the data set, but also improved the ability of the model to capture complex clinical patterns. As a result, the proposed models deliver more accurate and clinically relevant predictions, reinforcing their utility in guiding ICU management decisions for patients with aplastic anemia and offering a significant step forward in precision critical care.

\subsection{SHAP Analysis}\label{subsec:SHAP Analysis}

SHAP is a game-theoretic approach to interpreting machine learning models. It assigns each characteristic an importance score based on its contribution to the prediction of the model, ensuring a consistent and fair attribution of predictive power. SHAP values help us to understand not only which features influence predictions, but also the direction of their effect. In this study, SHAP analysis was used to examine the impact of various clinical variables on 7-day, 14-day, and 28-day mortality predictions using logistic regression models.

Across the performance of these SHAP analyses, APS III, bicarbonate, and platelet count consistently emerged as key predictors. APS III, a widely used ICU severity score, was of high importance, while bicarbonate levels reflected metabolic disturbances linked to poor outcomes. Platelet count was crucial in patients with aplastic anemia, as thrombocytopenia increases the risk of bleeding and mortality. The recurrence of these variables across multiple time horizons enhances the robustness of the model by demonstrating stability in the face of temporal variation and reinforces the biological plausibility of the predictors selected.

The importance of features varied between models. In the 7-day model, bicarbonate and platelet count dominated, highlighting acute metabolic and hematological effects on short-term survival. In the 14-day and 28-day models, APS III became more influential, indicating the growing role of systemic severity as patients' conditions evolved. Furthermore, dynamic variables such as lactate dehydrogenase and anion gap fluctuated in importance, reflecting changing metabolic and inflammatory states during the course of the ICU. The impact of the changes in characteristics illustrates the adaptability of the model to evolving clinical conditions, which is critical for accurate longitudinal risk stratification.

These results, which are shown in the study, suggest that while some prognostic factors remain consistently critical, their relative influence changes over time. Continuous monitoring and dynamic risk assessment are therefore essential for optimizing care in aplastic anemia patients in the ICU. By providing both global and local interpretability, SHAP analysis not only validates the clinical relevance of logistic regression models, but also improves clinician trust in predictive outputs. This combination of statistical rigor and clinical transparency supports the use of SHAP-augmented models for data-driven decision-making and the design of personalized treatment strategies tailored to individual risk trajectories.

\begin{figure}[H]
\centering
\includegraphics[width=90mm]{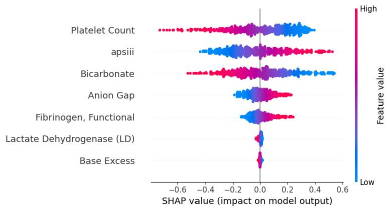}
\caption{(SHAP summary plot for the 7-day mortality prediction model.}
\label{fig:7d_shap}
\end{figure}

\begin{figure}[H]
\centering
\includegraphics[width=90mm]{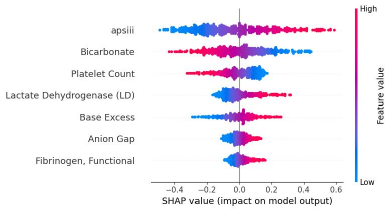}
\caption{SHAP summary plot for the 14-day mortality prediction model.}
\label{fig:14d_shap}
\end{figure}

\begin{figure}[H]
\centering
\includegraphics[width=90mm]{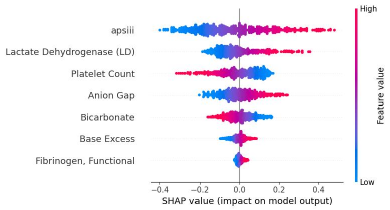}
\caption{SHAP summary plot for the 28-day mortality prediction model.}
\label{fig:28d_shap}
\end{figure}

\subsection{Permutation Importance Analysis}\label{subsec:Permutation Importance}

The importance of permutation is a widely used technique to assess the contribution of individual characteristics in predictive models. The method involves randomly shuffling the values of a single feature while keeping others unchanged and measuring the corresponding decrease in model performance. The importance of the characteristic \( X_j \) is calculated by comparing the baseline AUROC \( AUC_{\text{base}} \) with the AUROC after shuffling \( AUC_{\text{perm}} \):

\begin{eqnarray}
I(X_j) = AUC_{\text{base}} - AUC_{\text{perm}}(X_j)
\end{eqnarray}

where a larger \( I(X_j) \) indicates that the feature has a stronger influence on the predictive performance of the model. This process is repeated multiple times to obtain a robust estimate of the importance of the feature.

Permutation importance analysis was performed to evaluate the relative contribution of each feature to the performance of the model. Table~\ref{table:permutation_feature_importance} presents the main characteristics ranked by their impact on AUROC on the 7-day, 14-day, and 28-day mortality prediction models.

\begin{table}[H]
\caption{Significant permutation feature importance.}
\label{table:permutation_feature_importance}
\centering
\setlength{\tabcolsep}{6pt}
\begin{tabular}{lccc}
\noalign{\global\arrayrulewidth=0.8pt}\arrayrulecolor{black}\hline
\rowcolor{EDEDE9}
\textbf{Feature} & \textbf{AUC Drop (7d)} & \textbf{AUC Drop (14d)} & \textbf{AUC Drop (28d)} \\
\noalign{\global\arrayrulewidth=0.4pt}\hline
\rowcolor{E3D5CA}
\textbf{APS III} & \textbf{0.101} & \textbf{0.143} & \textbf{0.117} \\
Platelet Count & 0.045 & 0.037 & 0.038 \\
Bicarbonate & 0.042 & 0.039 & 0.011 \\
Anion Gap & 0.021 & 0.017 & 0.044 \\
\hline
\end{tabular}
\end{table}

APS III emerged as the most critical feature, with the largest AUC drops observed when it was permuted: 0.101 at 7 days, 0.143 at 14 days and 0.117 at 28 days. This consistent pattern highlights the dominant role of systemic severity, as captured by APS III, in the short-term mortality risk among ICU patients with aplastic anemia. The robustness of APS III across all time horizons reinforces the stability of the model and its reliance on clinically significant predictors.

Platelet count and bicarbonate levels also demonstrated substantial contributions to model performance. Platelet count showed moderate and consistent AUC declines in all prediction windows, reflecting the well-known clinical significance of thrombocytopenia in aplastic anemia. Similarly, bicarbonate, a key metabolic parameter, showed a notable impact, especially in the 7-day and 14-day models, suggesting that early metabolic disturbances are crucial determinants of short-term outcomes.

The anions gap exhibited smaller but still significant AUC drops, particularly at the 28-day horizon, indicating its role in capturing late-stage metabolic derangements and systemic stress responses. 

In general, the results of the importance of the permutation corroborate the findings of the SHAP analysis and the feature selection studies, further validating the clinical interpretability and robustness of the model. Using stable and pathophysiologically relevant characteristics such as APS III, platelet count, and bicarbonate, the model not only demonstrates strong predictive capacity, but also aligns with established medical knowledge, enhancing its potential.

\subsection{Nomograms}\label{Nomograms}

\begin{figure}[H]
\centering
\includegraphics[width=100mm]{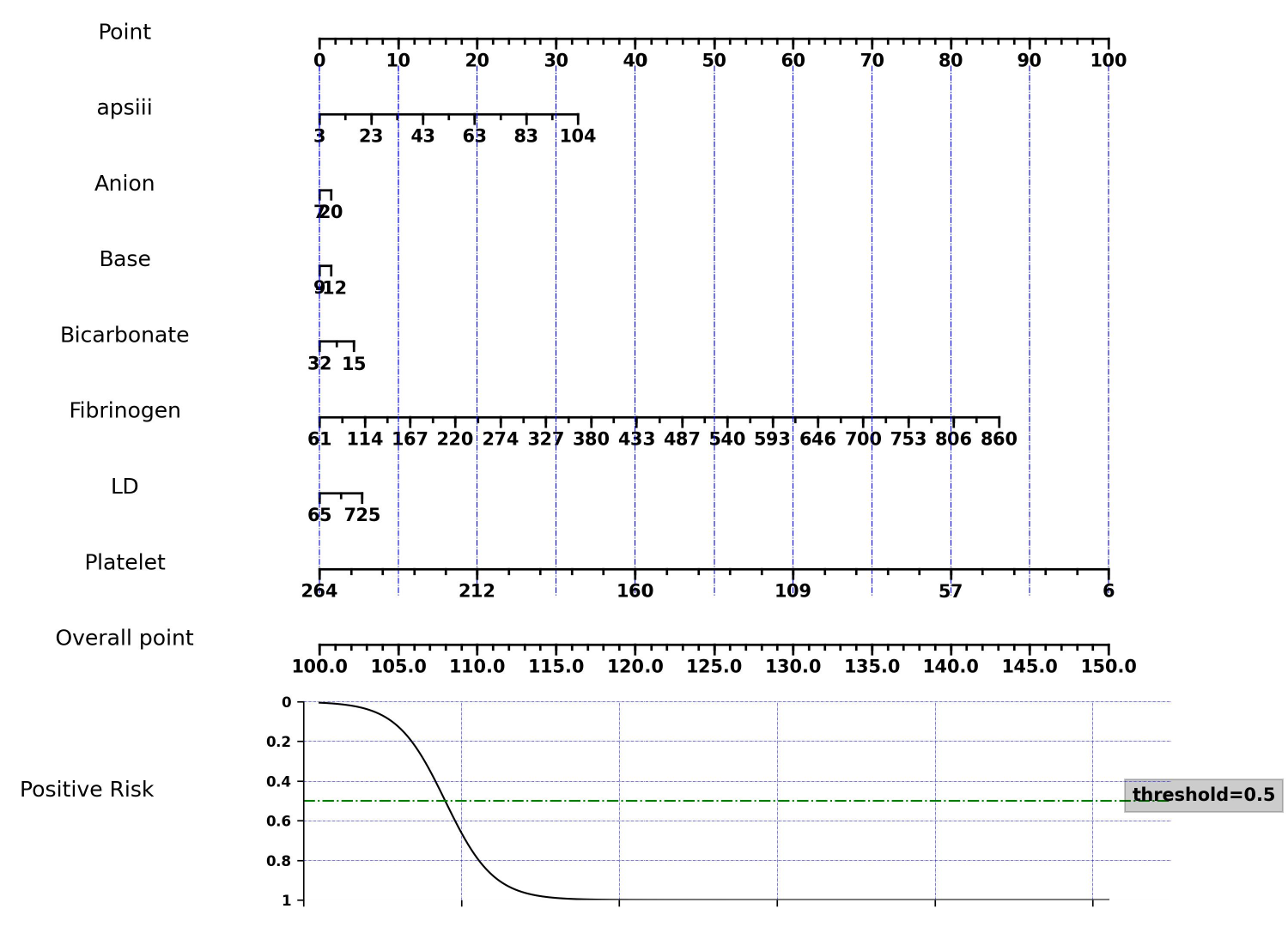}
\caption{Nomogram of 7 days survival prediction}
\label{fig:nomo_7}
\end{figure}

\begin{figure}[H]
\centering
\includegraphics[width=100mm]{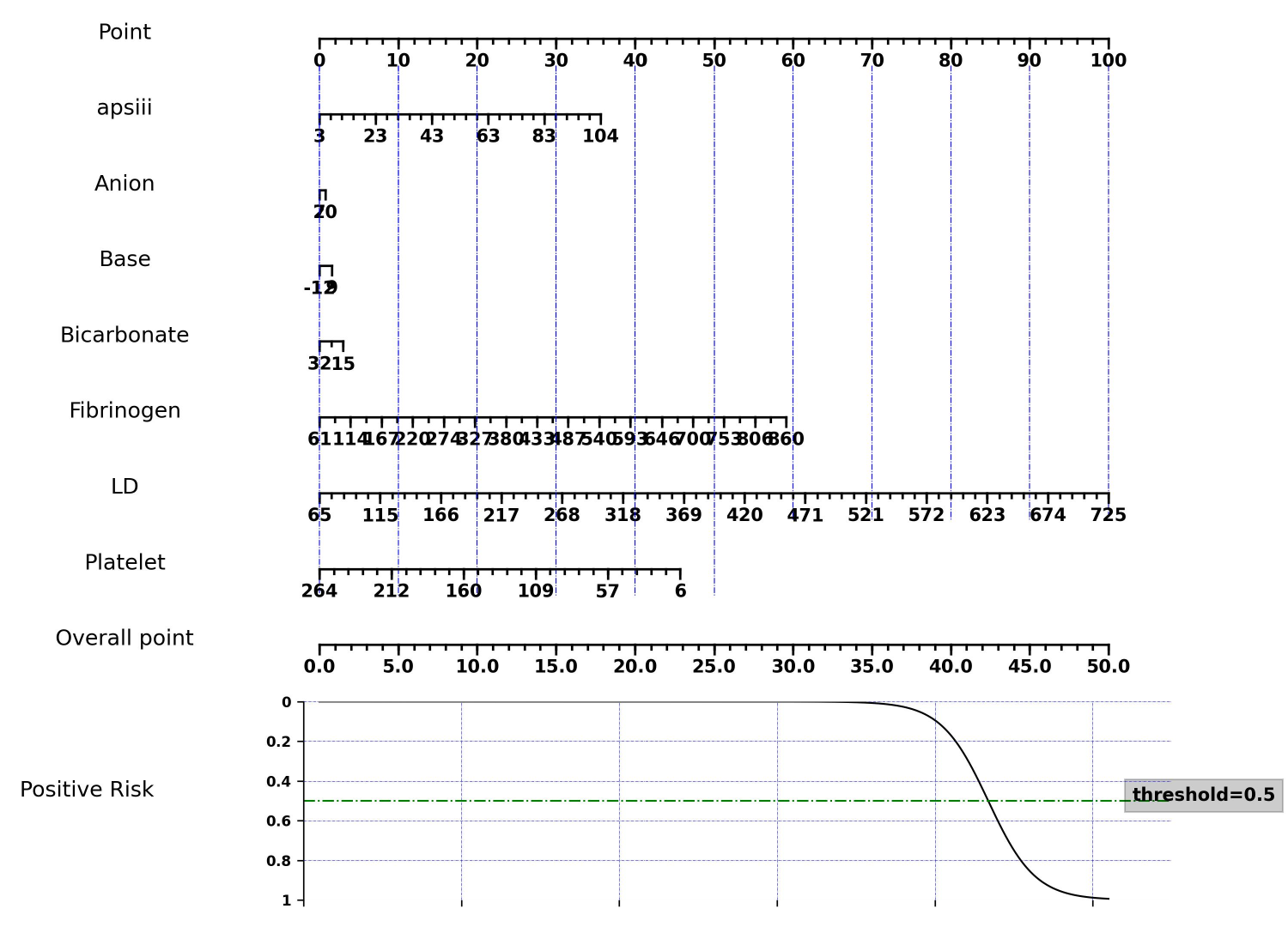}
\caption{Nomogram of 14 days survival prediction}
\label{fig:nomo_14}
\end{figure}

\begin{figure}[H]
\centering
\includegraphics[width=100mm]{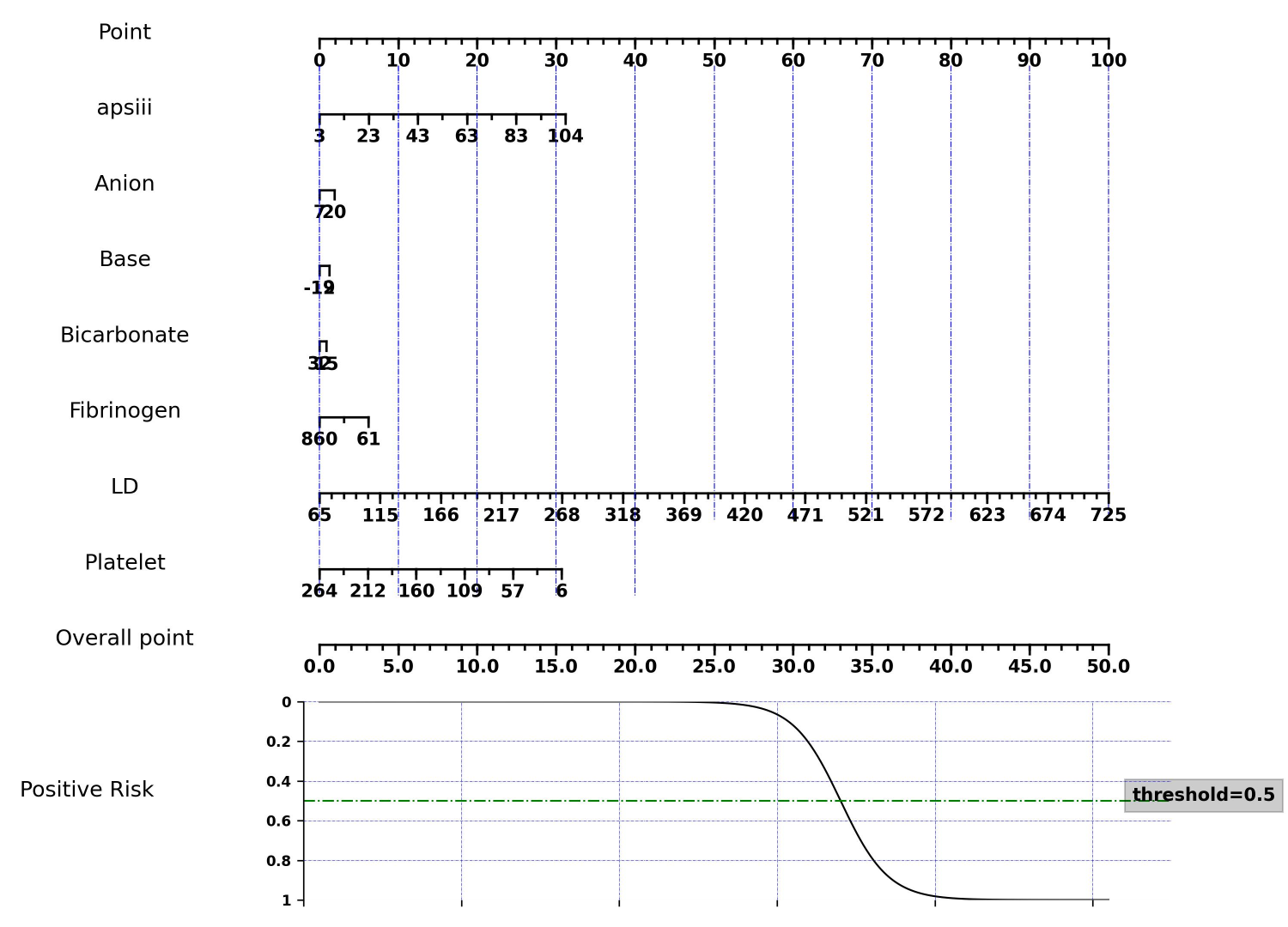}
\caption{Nomogram of 28 days survival prediction}
\label{fig:nomo_28}
\end{figure}

The nomograms, presented in Fig~\ref{fig:nomo_7}, Fig~\ref{fig:nomo_14}, and Fig~\ref{fig:nomo_28}, provide a dynamic and visual representation of mortality risk estimation in ICU patients with aplastic anemia over 7-, 14-, and 28-day intervals. Each nomogram integrates the seven selected clinical predictors—APS III, Anion Gap, Base Excess, Bicarbonate, Fibrinogen, Lactate Dehydrogenase (LD), and Platelet Count—into a scoring system that translates raw clinical measurements into an individualized probability of death.

Across all time windows, APS III consistently occupies the highest point allocation range, reaffirming its role as the most influential variable in the short-term mortality prediction. In the 7-day nomogram (Fig~\ref{fig:nomo_7}), the cumulative point total remains relatively low, reflecting the model's prioritization of acute physiological derangements such as bicarbonate fluctuations and thrombocytopenia. Base Excess also demonstrates a moderate but tangible contribution, suggesting that metabolic compensation plays a key role in ultra-early mortality.

In contrast, the 14-day nomogram (Fig~\ref{fig:nomo_14}) exhibits a broader spread in the contribution of both systemic and biochemical markers. APS III remains dominant, but fibrinogen and LD begin to show increased weight, indicating that inflammatory burden and tissue turnover become progressively relevant in the subacute phase of ICU stay. Notably, Bicarbonate maintains its importance, reflecting the ongoing impact of acid-base disturbances beyond the acute phase.

The 28-day nomogram (Fig~\ref{fig:nomo_28}) displays the most expansive point range, emphasizing the cumulative burden of systemic dysfunction. The scoring curve for LD and fibrinogen becomes steeper, underscoring their prognostic relevance in longer-term ICU outcomes, likely reflecting persistent cellular stress and coagulopathy. Anion Gap gains incremental influence, potentially signaling delayed-onset metabolic disturbances or secondary complications such as sepsis or renal dysfunction.

Despite their shared structural format, the three nomograms differ in score calibration, variable weighting, and risk thresholds. This temporal stratification enables clinicians to not only estimate current mortality risk but also anticipate evolving physiological trajectories. The inclusion of both static (e.g., APS III) and dynamic (e.g., bicarbonate, LD) indicators enhances their clinical utility in personalized ICU management.

\subsection{Interactive Nomogram}\label{Interactive Nomogram}
An interactive nomogram was developed to predict ICU mortality for patients with aplastic anemia at 7, 14 and 28 days. 

Patient data was extracted from the MIMIC-IV database for processing. Logistic regression was used to estimate the probability of mortality. An interactive web-based tool was developed using Dash and plotly to provide a user-friendly interface. A gauge-style nomogram was implemented to enable real-time risk assessment and visualization.

It is publicly available on GitHub at the following link:
\url{https://github.com/JunyiTim/Interactive-nomogram-for-ICU-patients-with-aplastic-anemia-}

\section{Discussion}
\subsection{Summary of existing model compilation}
This study introduces a clinically interpretable and high-performing predictive framework for short-term survival in ICU patients with AA. Through an integrative approach that combines clinical domain knowledge, advanced feature selection, and model interpretability, we developed three logistic regression-based nomograms to predict 7-day, 14-day, and 28-day mortality. The final models achieved AUROCs of 0.823, 0.831, and 0.830, respectively, on the internal MIMIC-IV test set, and maintained consistent performance on the external eICU validation set (AUROCs ranging from 0.709 to 0.739), demonstrating strong generalizability.

Key innovations include a two-step hybrid feature selection method that distilled over 400 candidate variables to 7 biologically meaningful predictors. To address the rare-event nature of early ICU mortality in AA, a novel custom SMOTE algorithm was introduced to enhance sampling in minority outcome strata. Furthermore, SHAP and permutation importance methods provided complementary interpretability, confirming the dominant role of APS III, platelet count, and bicarbonate in early mortality risk. An interactive Dash-based application further translated these findings into a real-time clinical tool.

Clinically, the model offers a decision support mechanism for ICU practitioners managing patients with this rare and high-risk hematologic disorder. By identifying high-risk patients early during ICU stay, the tool enables more precise triage, resource prioritization, and patient-family communication, supporting ICU operational efficiency and tailored therapeutic planning.

\subsection{Comparison with Prior Studies}
While multiple studies have explored mortality in patients with AA, few have addressed short-term ICU outcomes using multivariate, data-driven modeling frameworks. The majority of existing literature focuses on long-term survival after specific treatments, such as immunosuppressive therapy or HSCT, and lacks real-time ICU applicability.

Scheinberg et al. (2011) reported a 5-year survival rate of 89\% among early responders to horse ATG and cyclosporine therapy, versus 60\% in non-responders. Their findings emphasize the predictive value of early hematologic response, but they did not develop a risk scoring system, nor did they explore ICU-specific physiological predictors. Their work offers limited utility in the critical care setting where real-time decisions are required.

Wang et al. (2020) introduced telomere length as a predictive biomarker in AA patients undergoing unrelated donor HSCT. Patients with short leukocyte telomeres (<10th percentile) had significantly higher mortality post-transplant (HR = 1.78, 95\% CI: 1.18–2.69, P = 0.006). However, telomere assays are not widely available in most ICU environments, and the focus on transplant recipients limits the generalizability of their findings to broader ICU AA populations.

Liu et al.\ (2023) proposed an Early Death Risk Score derived from a cohort of 377 patients with very severe AA receiving immunosuppressive therapy. Their score achieved an AUROC of 0.862 in predicting early death and identified four risk factors: age $>$ 24, ANC $\leq 0.015 \times 10^9$/L, ferritin $>$ 900~\text{ng/mL}, and recurrent fevers. While their model demonstrated excellent discrimination, it was restricted to VSAA cases and non-ICU populations. Moreover, their tool lacks external validation and does not incorporate dynamic ICU variables such as APS III or acid-base status.

Tu et al. developed a Cox regression-based model for survival prediction in critically ill AA patients and reported a C-index of 0.643. In contrast, our Cox regression model achieved a significantly higher C-index of 0.761, reflecting an 18.3\% improvement in discriminatory ability. This performance gain is attributable to our comprehensive feature engineering, KNN-based imputation, and imbalanced data handling strategies, all of which were not present in Tu’s study.

Overall, our work advances the literature in several key dimensions: (1) it extends beyond post-treatment and long-term survival to address short-term ICU mortality; (2) it incorporates a broader and more heterogeneous ICU AA cohort; (3) it achieves external validation using multicenter ICU data; (4) it provides a practical, interpretable, and interactive tool ready for bedside deployment. No prior study simultaneously achieves these goals, making our model a significant step toward real-world clinical integration for high-risk AA patients in critical care.

\subsection{Limitations and Future Work}
Despite promising results, this study has several limitations. First, the rarity of aplastic anemia limited the cohort size, which constrained the use of more complex deep learning architectures and hindered subgroup analyses (e.g., stratified by treatment modality or HSCT eligibility). Second, although we included external validation using the eICU database, both datasets are based in the United States and may not reflect global practice patterns or demographic distributions. Third, treatment variables such as immunosuppressant dosing, transfusion burden, or antimicrobial regimens were not fully incorporated, which may influence short-term survival. Finally, the current model is static and does not yet support time-updated prediction during ongoing ICU care.

Future directions include validating the model prospectively in international ICU settings, incorporating longitudinal EHR and physiological monitoring data for dynamic risk prediction, and expanding the model to include treatment-specific and genomic features. Furthermore, integration with electronic health records and ICU clinical decision support systems would enable real-time deployment and potential closed-loop interventions. From a methodological standpoint, incorporating causal inference frameworks and temporal models (e.g., time-series transformers) could enhance the robustness and adaptability of survival predictions over time.

\section{Conclusion}
AA presents a critical challenge in intensive care settings due to its high mortality risk and limited ICU-specific prognostic tools. Personalized and timely survival prediction is crucial to inform clinical decisions, resource allocation, and family counseling. This study addressed that need by developing interpretable and clinically actionable models to predict 7-day, 14-day, and 28-day mortality in ICU patients with AA. Our results contribute meaningfully to the ongoing effort to integrate precision medicine approaches into critical care for rare hematologic disorders.

We used a structured and reproducible machine learning pipeline that utilizes the MIMIC-IV and eICU databases. The feature selection strategy combined filter-based (SelectKBest) and wrapper-based (RFE) methods, further guided by domain expertise to ensure biological plausibility. A custom SMOTE technique was introduced to handle severe class imbalance associated with short-term mortality outcomes. Logistic regression was selected as the final modeling approach due to its superior balance between discrimination and interpretability, and was embedded into both static and interactive nomogram tools for clinical use.

The final models achieved high and consistent performance across time windows, with AUROCs of 0.823 (7-day), 0.831 (14-day), and 0.830 (28-day) on the internal test set, and retained strong generalizability on the external eICU validation cohort (AUROC 0.709–0.739). Compared with prior studies—such as Liu et al.’s early death risk score (AUROC = 0.862 in non-ICU VSAA patients) and Tu et al.’s Cox model (C-index = 0.643)—our framework demonstrated improved accuracy, broader applicability, and enhanced clinical readiness via an interactive prediction interface.

The proposed nomogram-based models hold strong potential for implementation in ICU decision support systems. By integrating interpretable machine learning with routinely collected EHR variables, our approach offers a scalable and transparent solution for survival prediction in critically ill AA patients. Future studies should validate the model prospectively in international cohorts, explore dynamic time-updated predictions, and incorporate treatment-specific, genetic, and physiologic waveform data to further personalize risk estimation.

\section*{Declarations}

\textbf{Funding}

Not applicable.

\textbf{Conflict of interest/Competing interests}

The authors declare that the research was conducted in the absence of any commercial or financial relationships that could be construed as a potential conflict of interest.

\textbf{Ethics approval and consent to participate}

This study used deidentified, publicly available databases (MIMIC-IV and eICU Collaborative Research Database) and does not require institutional review board (IRB) approval or patient consent.

\textbf{Consent for publication}

Not applicable.

\textbf{Data availability}

The datasets analyzed in this study are publicly available from PhysioNet:

- MIMIC-IV: \url{https://physionet.org/content/mimiciv/}
- eICU Collaborative Research Database: \url{https://physionet.org/content/eicu-crd/}

\textbf{Materials availability}

Not applicable.

\textbf{Code availability}

The interactive nomogram developed in this study is available at: \url{https://github.com/JunyiTim/Interactive-nomogram-for-ICU-patients-with-aplastic-anemia}

\textbf{Author contributions}

JF conceived the study and led the experimental design and implementation. SC, LS, and YS contributed equally to data preprocessing, modeling, and manuscript drafting. EP, KA, GP, and MP provided clinical insights, medical interpretation, and guidance on the relevance of findings to patient care. MP supervised the entire project. All authors reviewed and approved the final version of the manuscript.






\bibliography{sn-bibliography}

\end{document}